\documentclass[runningheads]{llncs}
\usepackage{graphicx}

\usepackage{caption}
\usepackage{tikz}
\usepackage{comment}
\usepackage{amsmath,amssymb} 
\usepackage{color}
\usepackage[accsupp]{axessibility}  

\usepackage[width=122mm,left=12mm,paperwidth=146mm,height=193mm,top=12mm,paperheight=217mm]{geometry}
\usepackage{subcaption}
\usepackage[normalem]{ulem}
\usepackage{xspace}
\usepackage{amsfonts}

\usepackage{tikz}
\def\checkmark{\tikz\fill[scale=0.4](0,.35) -- (.25,0) -- (1,.7) -- (.25,.15) -- cycle;}
\usepackage{pifont}
\newcommand{\xmark}{\ding{55}}%
\usepackage{times}
\usepackage{epsfig}
\usepackage{graphicx}
\usepackage{amsmath, bm}
\usepackage{amssymb}
\usepackage{dsfont}
\usepackage{multirow}
\newcommand{\latinphrase}[1]{\textit{#1}} 
\newcommand{\etal}{\latinphrase{et~al.}\xspace}
\newcommand{\ie}{\latinphrase{i.e.}\xspace}

\begin{document}
\pagestyle{headings}
\mainmatter
\def\ECCVSubNumber{4193}  

\title{CrossFormer: Cross Spatio-Temporal Transformer for 3D Human Pose Estimation} 

\titlerunning{Abbreviated paper title}

\author{Mohammed Hassanin\inst{1} \and
Abdelwahed Khamiss\inst{2}\and
Mohammed Bennamoun\inst{3}\and
 Farid Boussaid\inst{3}\and
Ibrahim Radwan\inst{1}}

\authorrunning{M. Hassanin et al.}
%
\institute{Canberra University
\\ \email{d.mfawzy@gmail.com}\and
CSIRO\and
The University of Western Australia}

\maketitle
\begin{abstract}
3D human pose estimation can be handled by encoding the geometric dependencies between the body parts and enforcing the kinematic constraints. Recently, Transformer has been adopted to encode the long-range dependencies between the joints in the spatial and temporal domains. While they had shown excellence in long-range dependencies, studies have noted the need for improving the locality of vision Transformers. In this direction, we propose a novel pose estimation Transformer featuring rich representations of body joints critical for capturing subtle changes across frames (\textit{i.e.}, inter-feature representation). Specifically, through two novel interaction modules; Cross-Joint Interaction and Cross-Frame Interaction, the model explicitly encodes the local and global dependencies between the body joints. The proposed architecture achieved state-of-the-art performance on two popular 3D human pose estimation datasets, Human3.6 and MPI-INF-3DHP. In particular, our proposed CrossFormer method boosts performance by $0.9\%$ and $0.3\%$, compared to the closest counterpart, PoseFormer, using the detected 2D poses and ground-truth settings respectively.\footnote{Codes and models will be publicly available on \url{github.com}} 
\end{abstract}
%
\section{Introduction}
Automatic reconstruction of the 3D human pose from 2D images is a fundamental problem in computer vision. 3D human pose estimation solution provides a geometric representation that is important to many applications including human-computer interaction \cite{2021visual}, \cite{kisacanin2005real}, action understanding \cite{liu2019rgb}, \cite{hassanin2021learning}, healthcare \cite{ahad2019vision}, and motion analysis \cite{ahmedt2019vision}. The recently developed solutions to this problem can be categorised into two main groups:\textbf{(1)} Two-stage approaches such as \cite{zhao2019semantic} and \cite{cai2019exploiting}, where the input is firstly extracted using 2D human pose estimation (HPE) architectures (for instance, \cite{chen2018cascaded}, \cite{he2017mask}, \cite{radwan2019hierarchical}), and \textbf{(2)} End-to-end methods \cite{pavlakos2018ordinal}, \cite{moon2020i2l}, where 3D reconstruction is inferred directly from input images or videos. Owing to the recent advances in the area of 2D pose detectors, the two-stage approaches currently outperform their end-to-end counterparts.  

Despite research spanning decades, 3D human pose estimation remains very challenging. It is an ill-posed problem caused by the ambiguity and high degree of freedom in the input space \cite{radwan2013monocular}. Tackling the problem requires accounting the associated challenges. 2D to 3D pose regression is an under determined problem where many 3D poses might correspond to \textbf{almost identical} 2D projections. In this setting, even the slightest changes in joints positions and appearances can be informative. Given this, the problem calls for two natural requirements for successful regression. First, capturing rich per-joint feature representations to help mitigate the ambiguity and improve the accuracy (\texttt{R1:cross-feature interaction}). A potentially promising direction is leveraging cross-joints features correlations at a detailed level. Second, leveraging information across the Spatio-Temporal steam by attending carefully to features most relevant to the preformed pose (\texttt{R2:cross-frame interaction}). In this work, we draw on the recent advances of Vision Transformer and design an explicit mechanism for meeting those requirements.

\begin{figure}[t!]
    \centering
    \includegraphics[width=0.9\textwidth]{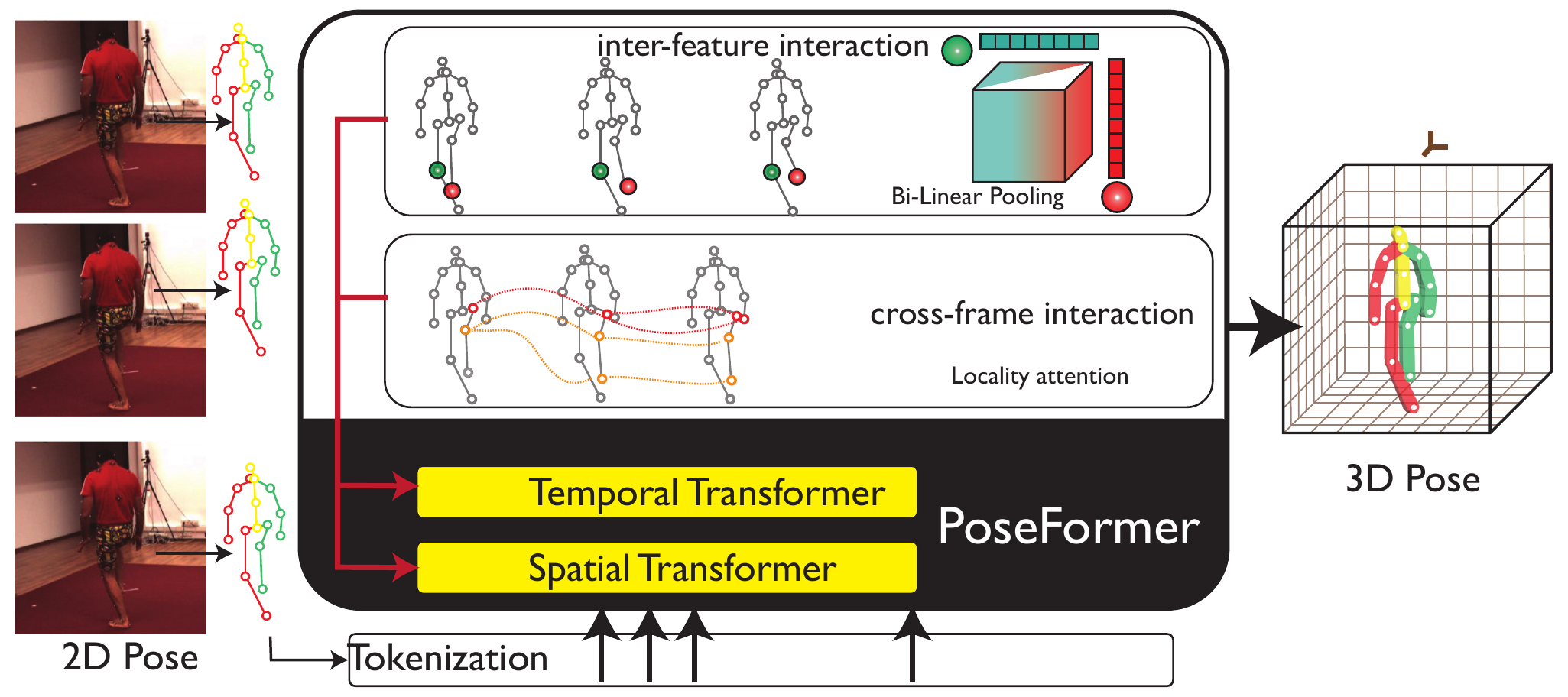}
    \caption{\textbf{Conceptual Illustration of CrossFromer Architecture}. This augments the PoseFormer\cite{poseformer}'s architecture with the proposed cross-feature (joints) and cross-frame interaction modules.}
    \label{fig:conceptual}
\vspace{-2em}
\end{figure}
%
The recent advent of transformers \cite{transformers} have progressed many visual recognition tasks. Transformers have been used to encode the long-range relationships between input tokens. As pose estimation is one of the fundamental computer vision problems, it has been approached by many Transformer-based architecture. Yet, one of most notable addition to this line of work is PoseFromer, \cite{poseformer}. Despite its great success, whose core is ViT \cite{dosovitskiy2020image}, it inherits some of the ViT limitations pointed out in the literature \cite{li2021localvit}, Namely, poor locality. This issue manifests itself in the fact that attention module attends to all tokens. While this design contributed to Transformer excellence in natural language domains, in vision applications it desirable attend to local information. As this limitation can be linked to \texttt{R2}, one can expect an improvement in PoseFormer by addressing it. More superficially, in this paper, we address the following question: \textit{is it possible to improve  Pose Transformers by improving locality and inter-feature ~representations}?. 

To answer the above question, we propose to integrate locality and rich inter-features interaction (as in Fig. \ref{fig:conceptual}), while retaining the key advantages of the original PoseFormer\cite{poseformer} (\textit{i.e.}, capacity to handle large number of tokens and Spatio-Temporal modelling). To this end, we design novel interaction modules to account for the above requirements as follows:
%

\texttt{R1)} To capture rich feature that highlights potentially feeble but effective details, we further integrate \textbf{Bi-linear Pooling} \cite{yue2018compact}  in the locality attention module by modifying the cross terms in the attention using outer product. Hence , expanding the attention to all channels (unlike the original inner product that merges information across channels dimension). Bi-linear Pooling was originally motivated by a similar goals of a fine-grained visual classification and has demonstrated success in many applications \cite{yu2021fast} from fine-grained recognition to  semantic segmentation and video classification. 

\texttt{R2)} There is a growing research that thrives for improving locality of Transformers using various approaches such as local attention \cite{ding2021ap} and regional attention maps \cite{chen2021regionvit}. Our work shares the same motivation, albeit using novel methodology. Inspired by non-local Neural Networks \cite{wang2018non}, \cite{yue2018compact}, we opt for \textbf{locality attention} \footnote{originally named ``non-local operation'' \cite{wang2018non} after the ``non-local'' mean operation \cite{buades2005non} and to set it apart for the local convolution. Here, we use ``locality attention''  to signify its role in our architecture and avoid confusion.} to leverage the feature representations of the joints across frames. This generalises the vanilla self-attention \cite{wang2018non}, module \cite{vaswani2017attention}, and can be interpreted as seeking a favourable middle ground between the locality-insensitive approaches (vanilla self-attention) and the purely local (stationary convolutional)~approaches.

To summarise, we combine the locality and inter-feature interaction in a transformer-based approach for 3D pose estimation. 
Our architecture realises the needed requirements and provides two novel cross interaction modules to encode both the local and global dependencies. More specifically, a cross-joint interaction (CJI) module is plugged in the spatial encoder of the Transformer architecture to encode the kinematic constraints between the body parts within a frame. This module (see Sec. \ref{sec:spatial}) is composed of depth-wise convolutions followed by group normalization and non-linearity layer (GELU). In addition to the cross-joint interaction module, we also propose a cross-frame interaction (CFI) module to handle interactions between the joints across frames. As opposed to PoseFormer \cite{poseformer}, where the inner-product is used to compute the correlation between frames, the CFI module explicitly learns the correlations between the frames by using the outer-product between feature representations across the frames. This helps in turn model the fine-grained temporal dynamics of the body parts. 

%
Experiments were performed on Human3.6 \cite{human} and MPI-INF-3DHP \cite{3DHP} datasets. Reported results demonstrate the superiority of proposed method over the state-of-the-art. Moreover, qualitative comparisons show that our method is efficient in capturing hardly-visible body parts. 

The main contributions of this paper are~:
\begin{itemize}
\item A cross-joint interaction module CJI for spatial transformer architectures to encode the kinematic dependencies between body joints while taking into account the local connections of each~joint.
\item A cross-frame interaction module CFI for temporal transformer architectures to capture the explicit correlations between body joints across frames.
\item State-of-the-art performance achieved on two popular benchmark datasets; Human3.6 and MPI-INF-3DHP.
\end{itemize}
\section{Related Work}
3D Human pose estimation methods are commonly used as the second stage to 2D detection methods. First, the input image is passed to detection frameworks \cite{he2017mask}, \cite{cpn} to infer 2D poses. Then, 2D poses are lifted to 3D using other methods \cite{poseformer}, \cite{pavllo20193d}. Martinez \etal \cite{martinez2017simple} used a fully-connected residual network to predict the 3D poses. Fang \etal \cite{fang2018learning} lifted to 3D poses using a grammar model for body joints configuration. Several other methods used  the temporal information to overcome the occlusion in the input images \cite{li2021exploiting}, \cite{pavllo20193d}. Pavllo \etal \cite{pavllo20193d} proposed a dilated temporal convolution approach to exploit the temporal information. Cai \etal \cite{cai2019exploiting} used a graph method to choose the center frame and then refine the final estimated 3D pose. Wang \etal \cite{wang2020motion} customized graph convolution network (GCN) in a U-shape as they involved motion modeling to learn the 3D poses. In \cite{zhao2019semantic}, a variant of a non-local module is customized to include the semantics of the input images.

Recently, vision transformers advanced all the visual recognition tasks \cite{transformers}. Following \cite{dosovitskiy2020image}, transformer has been used to lift 2D poses to the corresponding 3D poses. In \cite{lin2021end}, Lin \etal used convolutions and transformer together without temporal information to predict 3D poses. In order to eliminate the redundancy in the sequence with temporal information, Li \etal \cite{li2021exploiting} proposed a strided transformer network. In \cite{poseformer}, spatial-temporal transformer is used for 3D HPE tasks. Using transformers in HPE showed significant improvement overall. However, a pre-training on a large dataset is required to learn more representative and effective representations for the sequence HPE data. Our proposed method is different from the previous methods in leveraging the cross-interaction between the joints of the body parts in the spatial and the temporal domains. 
\section{Method}
\begin{figure}[t]
    \centering
    \includegraphics[width=0.98\textwidth]{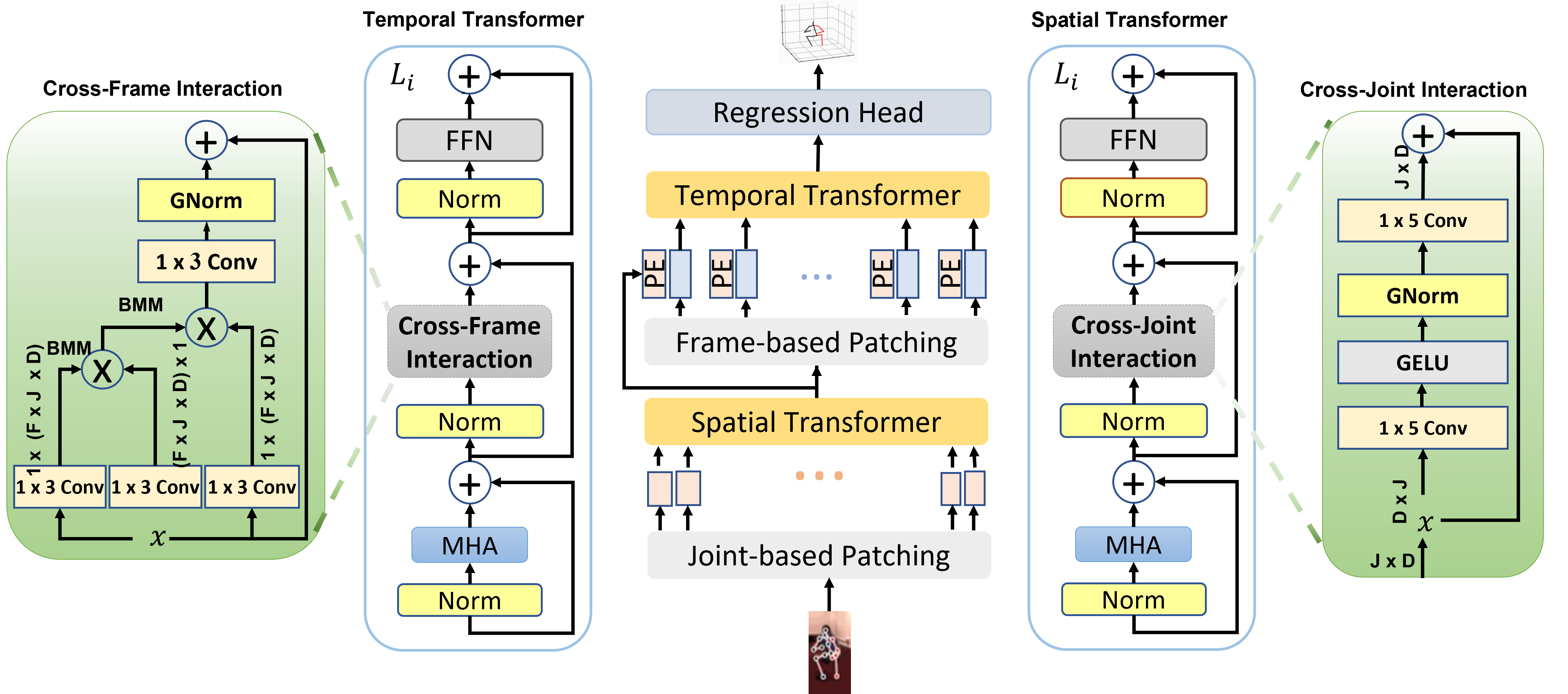}
    \caption{The proposed architecture is composed of two main modules: a spatial transformer along with the Cross-Joint Interaction module (CJI), and a temporal transformer with the proposed Cross-Frame Interaction (CFI).}
    \label{fig:method}
\vspace{-1em}
\end{figure}
This section presents the proposed architecture to estimate 3D human pose from 2D. Inspired by recently developed transformer approach, namely, Poseformer \cite{poseformer}, we propose interaction modules inside the spatial and temporal encoders to make the transformer more efficient when lifting the 2D to 3D poses. The 2D input poses can be inferred from any 2D pose detection approach such as \cite{chen2018cascaded}, \cite{he2017mask}. The poses of the consecutive frames in an input video are concatenated to form the input to the proposed architecture. Suppose, $\{\textbf{x}_1, \textbf{x}_2, ..., \textbf{x}_N\}$ denotes the set of the 2D input frames, where $\textbf{x}_i \in \mathbb{R}^{J \times 2}$ is composed of the 2D positions of the body joints for frame $i$, $N$ is the total number of frames in the input video and $J$ is the number of joints. The output of each frame is the 3D body joints, $\textbf{y} \in \mathbb{R}^{J \times 3}$. The proposed architecture incorporates cross-interaction modules into a vanilla spatial and temporal transformer \cite{dosovitskiy2020image}. Incorporating these modules with the transformer helps to capture both long-range relationships and local interactions between the body joints and across frames in both the spatial and temporal domains, respectively.

\subsection{Transformers}
Despite the great success achieved by transformers in computer vision tasks \cite{dosovitskiy2020image}, they focus mostly on global dependencies between frames in the input sequence (as observed by \cite{poseformer}). Motivated by this, we augment the proposed transformer with modules to capture more joints-related context information  within-frame  and corss-frames. Below we formulate our problem and review vision Transformer.

The input corresponds to the 2D poses in subsequent frames $\{\textbf{x}_{i} \in \mathbb{R}^{J \times 2} | i = 1, ..., N\}$. The initial layer of the transformer embeds the high dimensional features of each patch and their positional information. This step is called \textit{patch embedding}, and is achieved by the following projection operation:
\begin{equation}
\begin{split}
Z_0 &= [\textbf{x}_1 E; \textbf{x}_2 E; ..., \textbf{x}_P E] ,\\ 
&E \in \mathbb{R}^{(J\times2) \times D},  Z_0 \in \mathbb{R}^{P \times D}
\end{split}
\end{equation}
where $D$ is the embedding dimension, $P$ is the number of patches that is equivalent to the number of input body joints in the case of spatial processing and to the number of frames in the case of temporal processing. 

The output of the patch embedding step, $Z_0$, is then passed into the most important step of the transformer, \ie the \textit{self-attention}. It implicitly correlates the input patches in the form of attention scores. This step is mainly based on mapping three learnable weight matrices $Q, K, V \in \mathbb{R}^{P \times D}$ to attend for the output features. 

The self-attention operation is computed as a scaled dot-product between these matrices as follows:
\begin{equation}
    \mathit{A(Q, K, V)} = \mathrm{Softmax}(\frac{Q.K^T}{\sqrt{P}}).V
\end{equation}
This attention operation is applied through a multi-head attention (MHA), which combines various representations with different positions in parallel. The multi-head attention operation is simply achieved by concatenating all heads as~follows:
\begin{equation}
    \mathit{MHA} = \mathit{Concat}(A_j(.)).W, \quad j \in 1, ..., H
\end{equation}
where $W$ is a learnable weight matrix and $H$ is the number of heads.

The self-attention module is then combined with other layers such as layer normalisation \cite{wu2018group} and multi-layer perceptron (MLP). The steps in a transformer layer ($l$), which are following the patch embedding step can be listed, in general, as~follows:
\begin{equation}
    \begin{split}
        &Z_l = \mathit{MHA(LN(}Z_{l-1})) + Z_{l-1}, \\
        &Z_l = \mathit{MLP(LN(}Z_l))  + Z_l,\\
        &Z_l = \mathit{LN(}Z_l), \\
        &\mathit{where}\quad l = 1, 2, ..., L 
    \end{split}
    \label{eq:encoder}
\end{equation}
where $LN(.)$ represents the layer normalization and $L$ is the layer indicator of the transformer. Both the spatial and temporal parts of the transformer consist of identical layers. However, the output feature space for the spatial encoders $\in \mathbb{R}^{J \times D}$ and for the temporal encoders is in $\mathbb{R}^{F \times D}$, where F is number of frames in the input patch. The final layer is another linear projection step that maps the output space to $\mathbb{R}^{J\times3}$ for each frame.

\subsection{Spatial Interaction}
\label{sec:spatial}
The spatial encoders of the transformer learn the relationships between the body joints within the frame. The input is represented as $\textbf{x} \in \mathbb{R}^{J \times 2}$, where each joint is considered as an independent patch, and the output is the feature representation of each joint with respect to the other joints. Inspired by \cite{poseformer}, the 2D coordinates of each joint are firstly transformed using a linear operation. The output of this step, $Z_0 \in \mathbb{R}^{P \times D}$ is then passed forward to compute the self-attention scores. This encodes the dependencies between the different joints. However, these operations disregard the low-score relationships. This is due to the non-local nature of the transformer operations listed in Eq. \ref{eq:encoder}. 

In order to tackle this issue, we propose the Cross-Joints Interaction (CJI) module that we integrate inside the spatial encoders with an aim to achieve two characteristics; \textbf{1)} getting the transformer to consider the locality nature of the human body parts and their local interactions the same as encoding their non-local interactions (\ie long-range dependencies), and \textbf{2)} explicitly encoding the interaction between the joints of the body parts across the channels, which enriches the representation of the joints with low attention scores. Achieving these two characteristics improves the scores of the Multi-Head Attention for the 3D human pose estimation~task.\\
\noindent
\\
\textbf{Cross-Joints Interaction (CJI) Module}\\
This module is inserted between the MHA layer and MLP for each block. The CJI module consists of two depth-wise convolutions with kernel size $5$, group normalization and non-linearity GELU. Also, the residual connection is added to the output of the module to avoid overfitting. The operations within the CJI module are performed using the sequence of the following operations on outpout of the patch embedding step $Z_0$:
\begin{equation}
    \begin{split}
        Z &= \mathit{CONV(GN(GELU(CONV(Z))))} + Z
    \end{split}
\end{equation}
where $GELU$ refers to the non-linear layer in \cite{gelu}, $CONV$ is the standard convolution layer with kernel $5$ and $GN$ indicates the group normalization used in \cite{wu2018group}. Since the focus of the CJI module is on the cross-interaction between the joints, the output of the MHA part in Eq. \ref{eq:encoder} has been transposed. That is, it becomes $Z_0 \in \mathbb{R}^{D \times P}$. The spatial encoders for a transformer layer $l$ can then be represented by the following list of operations:
\begin{equation}
    \begin{split}
        &Z_l = \mathit{MHA(LN}(Z_{l-1})) + Z_{l-1}, \\
        &Z_l = \mathit{CONV(GN(GELU(CONV}(Z_l))))+Z_l\\
        &Z_l = \mathit{MLP(LN}(Z_l))  + Z_l\\
        &Z_l = \mathit{LN}(Z_l), 
    \end{split}
    \label{eq:encoder_cij}
\end{equation}
\vspace{-1em}
\subsection{Temporal Interaction}
\label{sec:temporal}
In contrast to the spatial encoders, which encode the long-range dependencies between the joints within each frame, the temporal counterparts aim at learning rich representations across frames. These encoders are stacked on top of the spatial ones. Their input is firstly flattened from $\mathbb{R}^{J \times D}$ to $\mathbb{R}^{1 \times (J \times D)}$ leading to $\mathbb{R}^{F \times (J \times D)}$ for all frames in a patch. Similar to the vanilla transformers \cite{transformers}, the temporal positional information are added to the input space. Apart from that, the remaining settings including self-attention modules and MLP blocks are just as in the case of spatial encoders. The input size of the transformer is maintained all over the transformers layers, which is $\mathbb{R}^{F \times (J \times D)}$.\\
\noindent
\\


\textbf{Cross-Frames Interaction (CFI) Module}\\
The whole attention in temporal encoders is based on the dependencies between the channels (\ie, $\mathbb{R}^{f \times (J \times D)}$), where the explicit interaction between the frames is disregarded. Depending on the scores produced by SoftMax calculation,  joints with low scores might be ignored in the process. For example, partially visible or occluded joint won't be properly represented and reflected in the 3D space. In order to resolve this problem, we propose a Cross-Frame Interaction Module, CFI, to explicitly encode the relationships between the same joint across frames using Bi-Linear Pooling operations \cite{yue2018compact} instead of the traditional SoftMax. This helps in learning the correlations between the channels explicitly as in \cite{hu2018squeeze} and reflects the the kinematic constraints on the output space. 

Briefly, the Bi-Linear Pooling learns pairwise feature correlations using the outer product.
%
%
Each element of the correlation matrix $C_{ij} = \sum_F Z_{i} Z_{j}$ is a Bi-Linear product of the corresponding embedded features of frames $i$ and  $j$ and then it is sum-pooled, where $Z_{i} \in \mathbb{R}^{J \times D}$ is the input feature of frame $i$. More precisely, the input is transformed by combining the positional information with the frames where $Z \in \mathbb{R}^ {F \times (J \times D)}$ and then using convolutions we extract $K,\ Q,\ \mathrm{and}\ V$ such that:
\vspace{-1em}
\begin{equation}
    K = ZW_k, \quad Q=ZW_q, \quad V=ZW_v 
\end{equation}
The bilinear  matrix multiplication is performed between matrices $Q$, $K$ and $V$ as follows:
\vspace{-1em}
\begin{equation}
\begin{split}
    &C = K \otimes Q \quad \in \mathbb{R}^{F \times F}   \\
    &Z = C\otimes V \quad \in \mathbb{R}^{D \times F}   \\
    \end{split}
\end{equation}
where $\otimes$ refers to the bilinear pooling operation. Then, the output is added to the input after performing convolution and group normalization. Compared to self-attention modules in Equation \ref{eq:encoder}, CFI module uses a bilinear pooling to learn pairwise interactions between the same joint across different frames. This highlights the discriminativeness of each frame which leads to rich representation. For example, one frame will focus on the top joints of the body and another on the lower part, while CFI will focus on combining both parts. Finally, CFI is merged with temporal transformer between MHA layer and MLP blocks. The updated sequence of the temporal transformer operations for a layer $l$ can be listed as follows:
\begin{equation}
\small
    \begin{split}
        &Z_l = \mathit{MHA(LN}(Z_{l-1})) + Z_{l-1}, \\
        &K = Z_lW_k,  Q=Z_lW_q,  V=Z_lW_v\\
        &Z_l = \mathit{GN(CONV}(((K \otimes Q)\otimes V))) + Z_l \\
        &Z_l = \mathit{MLP(LN}(Z_l))  + Z_l\\
        &Z_l = \mathit{LN}(Z_l), 
    \end{split}
    \label{eq:temporaltransformer}
\end{equation}
The sequence of transformer encoders is combined in a compact form, which enables an end-to-end training. Moreover, following the vanilla transformers \cite{poseformer} on using the three embedding matrices with the input \ie, $Q, K,$ and $V$ allows CJI and CFI to serve as generalised modules, which can be plugged in many other transformer architectures for other various visual recognition tasks.
\subsection{Regression Head}
The spatial and temporal transformers are stacked together as in Fig. \ref{fig:method}, where their input is passed to the spatial encoders and then to the temporal ones. The output of the temporal transformer is $\mathbb{R}^{F \times (J \times D)}$, which requires to be reduced to $\mathbb{R}^{1 \times (J \times 3)}$. First, 1D convolution is applied as a weighted average for the frames to transform to $\mathbb{R}^{1 \times (J \times D)}$. Then, a linear layer is used to learn the 3D geometries from the $D$ dimension followed by the normlisation layer. The final output is the estimated 3D position for each joint $\mathbb{R}^{1 \times (J \times 3)}$.
\subsection{Loss function}
Following the recent work in \cite{pavllo20193d}, the MPJPE loss function is employed to optimise the parameters of the whole architecture:
\vspace{-1em}
\begin{equation}
    \mathcal{L} = \frac{1}{J}\sum_{k=1}^J \lVert g_k - p_k\rVert_2,
\end{equation}

where $g_k$ represents the ground-truth 3D joint position of joint $k$ and $p_k$ is the 3D output of the proposed architecture of the $k$-th joint.
\begin{table*}[t]
\centering
\caption{Comparison between our proposed method and the state-of-the art approaches for 3D human pose estimation. Mean Per Joint Position Error(MPJPE) is used to measure the mean error between the estimated 3D pose and the ground truth 3D pose on
Human3.6M under Protocols 1\&2 where 2D pose detection is used as input. The top shows results of Protocol 1 (MPJPE), whereas the bottom part shows the results
of Protocol 2 (P-MPJPE). $f$ refers to the number of frames used in each method, $\ast$ denotes that the input 2D pose detection method used is
the cascaded pyramid network (CPN), and $\dagger$ refers to a transformer-based model. (\textcolor{red}{Red}: best; \textcolor{blue}{Blue}: second best)}
\resizebox{\textwidth}{!}{\begin{tabular}{l|c|ccccccccccccccc|c}
\hline 
Protocol 1 &  & Dir.  & Disc.  & Eat.  & Greet  & Phone  & Photo  & Pose  & Purch.  & Sit  & SitD.  & Somke  & Wait  & WalkD.  & Walk  & WalkT.  & Average\tabularnewline
\hline 
\hline 
 Debral \etal \cite{dabral2018learning}& ECCV\textquoteright 18 & 44.8  & 50.4  & 44.7 & 49.0  & 52.9  & 61.4  & 43.5 & 45.5  & 63.1  & 87.3 & 51.7  & 48.5  & 52.2 & 37.6  & 41.9 & 52.1 \tabularnewline

 Cai \etal \cite{cai2019exploiting} $(f = 7)$ & ICCV\textquoteright 19 & 44.6  & 47.4  & 45.6  & 48.8  & 50.8  & 59.0 & 47.2  & 43.9 & 57.9  & 61.9  & 49.7  & 46.6 & 51.3 & 37.1  & 39.4  & 48.8\tabularnewline

 Pavllo \etal \cite{pavllo20193d} $(f = 243)$* & CVPR\textquoteright 19  & 45.2  & 46.7  & 43.3  & 45.6  & 48.1  & 55.1  & 44.6  & 44.3  & 57.3  & 65.8  & 47.1  & 44.0  & 49.0  & 32.8  & 33.9  & 46.8\tabularnewline

Lin \etal \cite{lin2019trajectory}$(f = 50)$ & BMVC\textquoteright 19  & 42.5  & 44.8 & 42.6  & 44.2  & 48.5  & 57.1  & 52.6  & 41.4  & 56.5 & 64.5 & 47.4  & 43.0 & 48.1  & 33.0  & 35.1  & 46.6\tabularnewline

 Yeh \etal \cite{yeh2019chirality}& NIPS\textquoteright 19  & 44.8  & 46.1 & 43.3  & 46.4  & 49.0  & 55.2  & 44.6  & 44.0  & 58.3  & 62.7  & 47.1 & 43.9  & 48.6  & 32.7 & 33.3  & 46.7\tabularnewline

Liu \etal \cite{liu2020attention} $(f = 243)*$& CVPR\textquoteright 20  & 41.8 & 44.8 & 41.1  & 44.9 & 47.4  & 54.1 & 43.4 & 42.2  & 56.2  & 63.6  & \textcolor{blue}{45.3}  & 43.5  & 45.3  & \textcolor{blue}{31.3}  & 32.2  & 45.1\tabularnewline

SRNet \cite{zeng2020srnet} * & ECCV\textquoteright 20  & 46.6  & 47.1  & 43.9  & \textcolor{blue}{41.6}  & \textcolor{blue}{45.8}  & 49.6  & 46.5  & \textcolor{red}{40.0}  &\textcolor{blue} {53.4}  & 61.1  & 46.1  & 42.6  & \textcolor{red}{43.1}  & 31.5  & 32.6  & 44.8\tabularnewline

 UGCN \cite{wang2020motion} (f = 96)& ECCV\textquoteright 20  & \textcolor{blue}{41.3}  & \textcolor{blue}{43.9}  & 44.0  & 42.2  & 48.0  & 57.1  & 42.2  & 43.2  & 57.3  & 61.3  & 47.0  & 43.5  & 47.0  & 32.6  & \textcolor{blue}{31.8}  & 45.6\tabularnewline

 METRO \cite{lin2021end} (f = 1) †  & CVPR\textquoteright 21 & - & - & - & - & - & - & - & - & - & - & - & - & - & - & - & 54.0\tabularnewline

PoseFormer (no PT) \cite{poseformer} (81) & ICCV'21 & 43.0  & 46.5  & 41.4  & 44.1  & 48.1 & 53.2  & 43.7  & 43.6 & 54.9  & 62.3  & 47.1  & 44.9  & 47.7  & 32.8 & 33.5  & 45.7\tabularnewline

Chen \etal \cite{chen2021anatomy} (f = 81)* & TCSVT\textquoteright 21  & 42.1  &\textcolor{red}{ 43.8}  & 41.0  & 43.8  & 46.1  & 53.5  & 42.4  & 43.1  & 53.9  & \textcolor{red}{60.5}  & 45.7  & \textcolor{blue}{42.1}  & 46.2  & 32.2  & 33.8  & 44.6\tabularnewline
PoseFormer (PT)\cite{poseformer} (81) & ICCV'21 & 41.5  & 44.8  & \textcolor{red}{39.8}  & 42.5  & 46.5 & \textcolor{red}{51.6}  & \textcolor{blue}{42.1}  & 42.0 & \textcolor{red}{53.3}  & \textcolor{blue}{60.7}  & 45.5  & 43.3  & 46.1  & 31.8 & 32.2  & \textcolor{blue}{44.3}\tabularnewline
\hline
CrossFormer (81) &  & \textcolor{red}{40.7} & 44.1  & \textcolor{blue}{40.8}  & \textcolor{red}{41.5}  & \textcolor{red}{45.8} & \textcolor{blue}{52.8}  & \textcolor{red}{41.2}  & \textcolor{blue}{40.8} & 55.3  & 61.9  & \textcolor{red}{44.9}  & \textcolor{red}{41.8}  & \textcolor{blue}{44.6}  & \textcolor{red}{29.2} & \textcolor{red}{31.1}  & \textcolor{red}{43.7}\tabularnewline
\hline

Protocol 2 &  & Dir.  & Disc.  & Eat.  & Greet  & Phone  & Photo  & Pose  & Purch.  & Sit  & SitD.  & Somke  & Wait  & WalkD.  & Walk  & WalkT.  & Average\tabularnewline
\hline
 Pavlakos \etal \cite{pavlakos2018ordinal}
& CVPR\textquoteright 18 & 34.7  & 39.8  & 41.8  & 38.6  & 42.5  & 47.5  & 38.0  & 36.6  & 50.7  & 56.8  & 42.6  & 39.6  & 43.9  & 32.1  & 36.5  & 41.8\tabularnewline

 Hossain \etal \cite{} 
& ECCV\textquoteright 18  & 35.7  & 39.3  & 44.6  & 43.0  & 47.2  & 54.0  & 38.3  & 37.5  & 51.6  & 61.3  & 46.5  & 41.4  & 47.3  & 34.2  & 39.4  & 44.1\tabularnewline

 Cai \etal \cite{cai2019exploiting} (f = 7) 
& ICCV\textquoteright 19  & 35.7  & 37.8  & 36.9  & 40.7  & 39.6  & 45.2  & 37.4  & 34.5 & 46.9  & 50.1  & 40.5  & 36.1  & 41.0  & 29.6  & 32.3  & 39.0\tabularnewline

Lin \etal \cite{lin2019trajectory} (f = 50)
 & BMVC\textquoteright 19  & 32.5  & 35.3  & 34.3  & 36.2  & 37.8  & 43.0  & 33.0  & 32.2  & 45.7  & 51.8  & 38.4  & 32.8  & 37.5  & 25.8  & 28.9  & 36.8\tabularnewline

Pavllo \etal \cite{pavllo20193d} (f = 243)* 
 & CVPR\textquoteright 19  & 34.1  & 36.1  & 34.4  & 37.2  & 36.4  & 42.2  & 34.4  & 33.6  & 45.0  & 52.5  & 37.4  & 33.8  & 37.8  & 25.6  & 27.3  & 36.5\tabularnewline

Liu \etal \cite{liu2020attention} (f = 243)* 
 & CVPR\textquoteright 20  &\textcolor{blue}{32.3}  & 35.2  & 33.3  & 35.8  & 35.9  & 41.5  & 33.2  & 32.7  & 44.6  & 50.9  & 37.0  & 32.4  & 37.0  & 25.2  & 27.2  & 35.6\tabularnewline

UGCN \cite{wang2020motion} (f = 96) 
 & ECCV\textquoteright 20 & 32.9  & 35.2  & 35.6  & \textcolor{blue}{34.4}  & 36.4  & 42.7  & \textcolor{red}{31.2}  & 32.5  & 45.6  & 50.2  & 37.3  & 32.8  & 36.3  & 26.0  & 23.9  & 35.5\tabularnewline

Chen \etal \cite{chen2021anatomy} (f = 81)* & TCSVT\textquoteright 21  & 33.1 & 35.3  & 33.4  & 35.9  & 36.1 & 41.7  & 32.8 & 33.3  & \textcolor{red}{42.6}  & \textcolor{blue}{49.4}  & 37.0  & 32.7  & 36.5  & 25.5  & 27.9 & 35.6\tabularnewline

PoseFormer (no PT) \cite{poseformer} (f=81) & ICCV'21 & 33.5  & 35.6  & 33.5 & 35.6 & 36.1  & 40.4  & 32.8  & 32.5 & 43.5  & 49.3  & \textcolor{blue}{35.4}  & 33.2 & 36.3  & 25.3 & 26.6  & 35.3\tabularnewline
\hline
PoseFormer \cite{poseformer}(f=81) & ICCV'21 & 32.5  & \textcolor{blue}{34.8}  & \textcolor{blue}{32.6} & 34.6 & \textcolor{blue}{35.3}  & \textcolor{red}{39.5}  & 32.1  & \textcolor{blue}{32.0} & \textcolor{blue}{42.8}  & \textcolor{red}{48.5}  & \textcolor{red}{34.8}  & \textcolor{blue}{32.4} & \textcolor{blue}{35.3}  & \textcolor{blue}{24.5} & \textcolor{blue}{26.0}  & \textcolor{blue}{34.6}\tabularnewline
\hline
CrossFormer (f=81) &  & \textcolor{red}{ 31.4} & \textcolor{red}{34.6}  & \textcolor{red}{32.6}  & \textcolor{red}{33.7}  & \textcolor{red}{34.3} & \textcolor{blue}{39.7}  & \textcolor{blue}{31.6}  & \textcolor{red}{31.0} & 44.3  & 49.3  & 35.9  & \textcolor{red}{31.3}  & \textcolor{red}{34.4}  & \textcolor{red}{ 23.4} & \textcolor{red}{25.5}  & \textcolor{red}{ 34.3}\tabularnewline\end{tabular}}
\label{table:h36}
\end{table*}
\noindent
\section{Experiments}
\label{sec:evaluation}
In this section, we provide empirical experiments to show the significance of our proposed method. First, we describe the used datasets, the evaluation criteria and protocols. Then, further experiments are conducted along with ablation studies. Finally, we provide the comparisons with state-of-the-art methods. 
%
\subsection{Datasets and Evaluation Protocols}
\textbf{Datasets}
Our experiments are evaluated on the most popular datasets for HPE tasks, Human3.6 \cite{human} and MPI-INF-3DHP \cite{3DHP}.\\
\textbf{Human3.6} dataset is the most popular dataset and the largest one for HPE. It consists of 3.6 million images in the form of video frames. It includes seven subjects and a total of 15 actions including ``walking'', ``sitting'' and ``posing''. Each video is captured from 4 different views. 3D annotations are provided by an accurate marker-based motion capture. The subjects are split for training and testing as in \cite{pavllo20193d}, where S1, S5, S6, S7and S8 are used for training and S1 and S11 for testing. One model is used to train all the frames for the various actions. All the videos are recorded in indoor scenes.\\
\textbf{MPI-INF-3DHP} dataset contains 8 actions from 14 different views which result to in diverse poses. It contains indoor and outdoor complex scenes and thus it is more challenging for HPE tasks. The test set includes 6 various subjects. Without stated, the settings are following \cite{pavlakos2018ordinal}. The scenarios of test set are: studio with a Green Screen (GS), studio without Green Screen (noGS) and outdoor scene (Outdoor).
\\
\textbf{Evaluation Protocols:}\\
\textbf{Human3.6} dataset evaluation for 3D pose estimation performance relative to the 3D ground-truth  follow \cite{pavllo20193d} using the most common metrics, \ie MPJPE and P-MPJPE. MPJPE (Mean Per Joint Position Error) refers to the average of Euclidean distance in millimeters between the predicted 3D human-body joints and the ground-truth ones. It is also denoted by Protocol 1. For P-MPJPE, it calculates the Euclidean distance between the 3D predicted pose after rigid alignment and the ground-truth. It is referred to Protocol 2 as it is more robust to prediction failure of the joint individuals.\\
\textbf{For MPI-INF-3DHP,} Protocol 1 is used, Area Under Curve (AUC),  and  Percentage of Correct Keypoint (PCK) within the 150mm as defined in \cite{3DHP} 
\begin{table}
\caption{Comparison between the estimated 3D pose of the proposed method and the ground truth 3D pose on
Human3.6M dataset using the Mean Per Joint Position Error under Protocol 1 (MPJPE). All methods use the ground truth 2D pose as input. (\textcolor{red}{Red}: best; \textcolor{blue}{Blue}: second best)}
\resizebox{\textwidth}{!}{\begin{tabular}{l|c|ccccccccccccccc|c}
\hline 
Protocol \#1 &  & Dir.  & Disc & Eat & Greet  & Phone  & Photo & Pose  & Purch.  & Sit  & SitD.  & Smoke  & Wait  & WalkD.  & Walk  & WalkT.  & Avg.\tabularnewline
\hline 

 Martinez \etal \cite{martinez2017simple}
& ICCV’17 & 37.7  & 44.4  & 40.3  & 42.1 & 48.2  & 54.9 & 44.4  & 42.1 & 54.6  & 58.0  & 45.1  & 46.4  & 47.6 & 36.4  & 40.4  & 45.5\tabularnewline

 Lee \etal \cite{lee2018propagating}
& ECCV’18 & 32.1  & 36.6 & 34.3  & 37.8  & 44.5 & 49.9  & 40.9  & 36.2  & 44.1 & 45.6 & 35.3 & 35.9  & 30.3  & 37.6 & 35.5  & 38.4\tabularnewline

Pavllo \etal \cite{pavllo20193d}
 & CVPR’19 & 35.2  & 40.2  & 32.7 & 35.7  & 38.2  & 45.5  & 40.6 & 36.1 & 48.8  & 47.3  & 37.8  & 39.7  & 38.7  & 27.8 & 29.5 & 37.8\tabularnewline

Cai \etal \cite{cai2019exploiting}
f = 243 & ICCV\textquoteright 19 & 32.9  & 38.7  & 32.9 & 37.0 & 37.3  & 44.8  & 38.7  & 36.1 & 41.0 & 45.6  & 36.8 & 37.7  & 37.7  & 29.5  & 31.6 & 37.2\tabularnewline

Xu \etal \cite{xu2021graph}
 &  CVPR’21 & 35.8  & 38.1 & 31.0 & 35.3  & 35.8  & 43.2  & 37.3 & 31.7 & 38.4 & 45.5  & 35.4 & 36.7  & 36.8 & 27.9  & 30.7  & 35.8\tabularnewline

 Liu \etal \cite{liu2020attention}
(f=243)& CVPR\textquoteright 20 & 34.5  & 37.1 & 33.6  & 34.2  & 32.9  & 37.1  & 39.6  & 35.8  & 40.7 & 41.4  & 33.0  & 33.8 & 33.0  & 26.6  & 26.9  & 34.7 \tabularnewline

Chen \etal \cite{chen2021anatomy}
(f=243)& TCSVT\textquoteright 21 & - & - & - & - & - & - & - & - & - & - & - & - & - & - & - & 32.3\tabularnewline

SRNet \cite{zeng2020srnet}    & ECCV\textquoteright 20 & 34.8  & \textcolor{blue}{32.1} & \textcolor{blue}{28.5} & \textcolor{blue}{30.7}  & 31.4  & 36.9 & 35.6 & \textcolor{blue}{30.5}  & 38.9  & 40.5 & 32.5  & \textcolor{blue}{31.0}  & 29.9  & \textcolor{blue}{22.5}  & 24.5  & 32.0\tabularnewline

PoseFormer \cite{poseformer} (f=81)& ICCV \textquoteright 21 & \textcolor{blue}{30.0}  & 33.6  & 29.9  & 31.0 & \textcolor{blue}{30.2}  & \textcolor{blue}{33.3}  & \textcolor{blue}{34.8}  & 31.4  & \textcolor{blue}{37.8} & \textcolor{blue}{38.6}  & \textcolor{blue}{31.7}  & 31.5  & \textcolor{blue}{29.0}  & 23.3  & \textcolor{blue}{23.1}  & \textcolor{blue}{31.3}\tabularnewline


\hline 
 CrossFormer (f=81)&  & \textcolor{red}{26.0}  &  \textcolor{red}{ 30.0} &\textcolor{red}{ 26.8}  &\textcolor{red}{ 26.2}  &\textcolor{red}{ 28.0}  & \textcolor{red}{ 31.0} &\textcolor{red}{ 30.4}  &\textcolor{red}{ 29.6}  & \textcolor{red}{ 35.4} & \textcolor{red}{ 37.1} &\textcolor{red}{ 28.4}  &\textcolor{red}{ 27.3}  &\textcolor{red}{ 26.7}  &\textcolor{red}{ 20.5}  &\textcolor{red}{ 19.9}  & \textcolor{red}{ 28.3}\tabularnewline
\hline 
\end{tabular}
}
\label{table:gt_h36}
\end{table}
\vspace{-2em}
\begin{figure}[t!] \  input   \quad\   GT \quad   Poseformer \ \ CrossFormer  \ input \   GT \   Poseformer \ \ CrossFormer
  \hrulefill
\resizebox{\textwidth}{!}{
\begin{subfigure}{0.13\paperwidth}
\includegraphics[width=0.13\paperwidth]{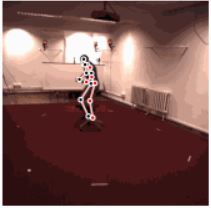}%
\end{subfigure}
\begin{subfigure}{0.17\paperwidth}
\includegraphics[width=0.17\paperwidth]{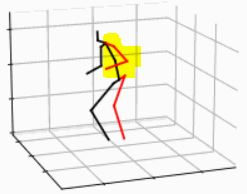}%
\end{subfigure}
\begin{subfigure}{0.18\paperwidth}
\includegraphics[width=0.18\paperwidth]{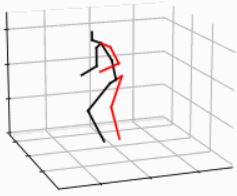}%
\end{subfigure}
\begin{subfigure}{0.17\paperwidth}
\includegraphics[width=0.17\paperwidth]{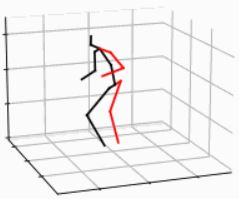}%
\end{subfigure}
\begin{subfigure}{0.13\paperwidth}
\includegraphics[width=0.13\paperwidth]{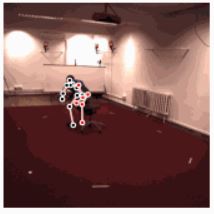}%
\end{subfigure}
\begin{subfigure}{0.17\paperwidth}
\includegraphics[width=0.17\paperwidth]{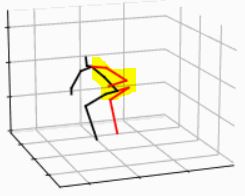}%
\end{subfigure}
\begin{subfigure}{0.18\paperwidth}
\includegraphics[width=0.18\paperwidth]{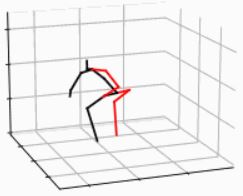}%
\end{subfigure}
\begin{subfigure}{0.17\paperwidth}
\includegraphics[width=0.17\paperwidth]{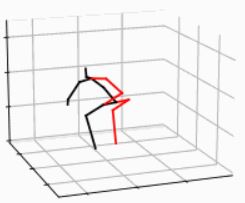}%
\end{subfigure}}
\resizebox{\textwidth}{!}{
\begin{subfigure}{0.13\paperwidth}
\includegraphics[width=0.13\paperwidth]{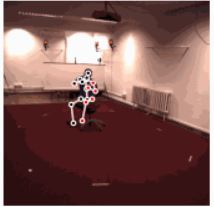}%
\end{subfigure}
\begin{subfigure}{0.17\textwidth}
\includegraphics[width=0.17\paperwidth]{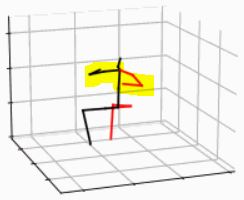}%
\end{subfigure}
\begin{subfigure}{0.18\paperwidth}
\includegraphics[width=0.18\paperwidth]{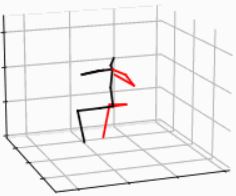}%
\end{subfigure}
\begin{subfigure}{0.17\paperwidth}
\includegraphics[width=0.17\paperwidth]{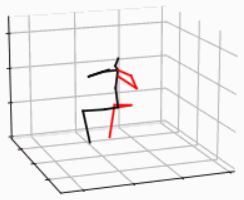}%
\end{subfigure}
\begin{subfigure}{0.13\paperwidth}
\includegraphics[width=0.13\paperwidth]{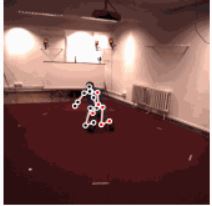}%
\end{subfigure}
\begin{subfigure}{0.17\paperwidth}
\includegraphics[width=0.17\paperwidth]{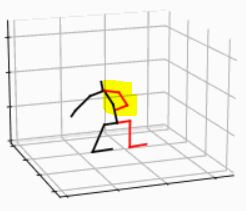}%
\end{subfigure}
\begin{subfigure}{0.18\paperwidth}
\includegraphics[width=0.18\paperwidth]{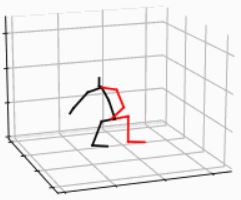}%
\end{subfigure}
\begin{subfigure}{0.17\paperwidth}
\includegraphics[width=0.17\paperwidth]{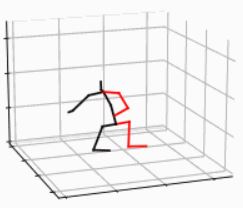}%
\end{subfigure}}
\resizebox{\textwidth}{!}{
\begin{subfigure}{0.13\paperwidth}
\includegraphics[width=0.13\paperwidth]{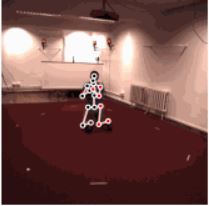}%
\end{subfigure}
\begin{subfigure}{0.17\paperwidth}
\includegraphics[width=0.17\paperwidth]{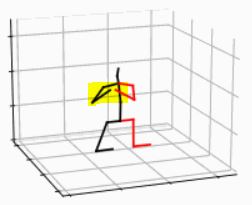}%
\end{subfigure}
\begin{subfigure}{0.18\paperwidth}
\includegraphics[width=0.18\paperwidth]{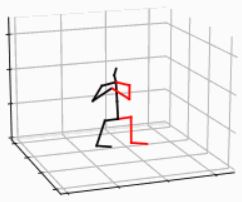}%
\end{subfigure}
\begin{subfigure}{0.17\paperwidth}
\includegraphics[width=0.17\paperwidth]{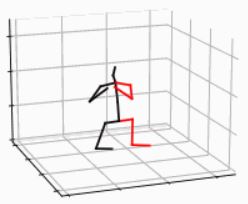}%
\end{subfigure}
\begin{subfigure}{0.13\paperwidth}
\includegraphics[width=0.13\paperwidth]{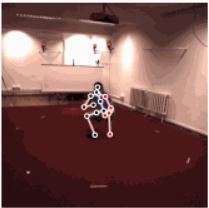}%
\end{subfigure}
\begin{subfigure}{0.17\paperwidth}
\includegraphics[width=0.17\paperwidth]{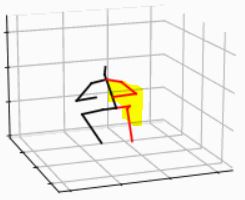}%
\end{subfigure}
\begin{subfigure}{0.18\paperwidth}
\includegraphics[width=0.18\paperwidth]{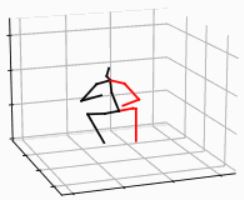}%
\end{subfigure}
\begin{subfigure}{0.17\paperwidth}
\includegraphics[width=0.17\paperwidth]{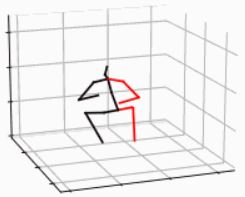}%
\end{subfigure}}
\caption{Visual qualitative comparison of the proposed method (CrossFormer), the ground-truth  and the state-of-the art approach (PoseFormer) \cite{poseformer}. The experiments are conducted on Human3.6M test set S11
with the ``SittingDown'' action. The blue arrows on the ground-truth highlight the locations where our method clearly behaves better.}
\label{fig:qualitative}%
\end{figure}

\subsection{Implementation Details}
We used Pytorch \cite{pytorch} to implement our proposed method. Two Tesla A100 GPU 40 GB each were used to run the experiments. Adam optimizer \cite{adam} is chosen to train the model for 100 epochs, decaying with $10\%$. Another decaying schedule is used with initial learning rate of $0.0001$ and an exponential decaying factor of $0.99$ after each epoch, whereas the batch size is $512$. We follow \cite{pavllo20193d} for the selection of the 2D pose detector, which is the cascaded pyramid network (CPN) \cite{cpn} on Human3.6 dataset. For MPI-INF-3DHP dataset, 2D ground-truth poses are used as in \cite{pavllo20193d}. Both of the spatial and temporal transformers have 4 layers, and the multi-head attention has 8 heads. The dimension of the features is 32 for spatial transformer and 544 for the temporal one. The receptive fields are 9, 27 and 81. Horizontal flip augmentation is used for the training and testing stages.
\subsection{Comparison with the state-of-the art}
\textbf{Human3.6} In this part, we compare the proposed method with the state-of-the art methods on Human3.6 dataset. 15 actions have been selected from two subjects, S9 and S11, for the evaluation.  In order to guarantee fair comparisons, the input is taken from CPN in the form of 2D keypoints for training and testing.  Table \ref{table:h36} shows the comparison of the SOTA methods with the proposed method (81 frames). Overall, our method achieves the state-of-the art on Human3.6 on all the metrics and it outperforms the state-of-the art (Chen \etal \cite{chen2021anatomy}) with a considerable margin of $0.9\%$, $1.3\%$ for Protocols 1 and 2, respectively. It is worth noting that the across-joint modules in the spatial and temporal cases are crucial to infer the body-joints dependencies. Comparing the proposed method with PoseFormer (with no pre-training used) shows  the significance of the across-joint correlation modules. Our method outperforms with a large margin of $2\%$ the SOTA. In terms of accuracy, we achieve $1\%$ better than the second best accuracy. Additionally, the proposed method achieves the best performance amongst all the compared methods in protocol 2 in Table \ref{table:h36} (bottom). In some selected difficult poses such as walk together, walk, smoke, where the poses change very quickly, the proposed method showed a significant improvement ranging from $1.1\%$ to  $2.5\%$ over the baseline. This highlights the ability of our method to encode the long-range interactions between the body-joints. Considering the pre-trained baseline, the proposed method achieves better performance for all the actions. These results show the importance of plugging the across-joints modules in the transformers.

Further experiments on Human3.6 using ground-truth 2D poses as input have also been performed. This shows the power of the proposed method where there is no noise in the input as in the previous case. Table \ref{table:gt_h36} shows the comparisons of our method and the baselines. Overall, the proposed method achieved the best performance amongst the baselines. It achieved $28.3\%$ MPJPE, whereas the second best approach achieves $31.3$ with gain of $3\%$. The proposed method outperforms the baselines in all the actions with a considerable improvement range from $2.4\%$ as the minimum difference, and $4.8\%$ for the largest.
\\
\textbf{MPI-INF-3DHP} 
We further compare the proposed methods (CrossFormer) to previous ones on MPP-INF-3DHP using 9 frames.  This is important because it illustrates the ability of the proposed method to train with fewer training samples in an outdoor settings. As Table \ref{tab:INF} shows, our method obtains the best performance amongst the compared ones \textit{w.r.t.} the metrics.
\begin{table}
\centering
\caption{Comparison between the proposed method (CrossFormer) and previous SOTA methods on MPI-INF-3DHP. The metrics of the comparison are the Percentage of Correct Keypoints (PCK) and Area Under the Curve (AUC). The best scores are marked in bold}
\begin{tabular}{l|l|ccc}
\hline 
 &  & PCK \textuparrow{}  & AUC \textuparrow{} & MPJPE \textdownarrow{}\tabularnewline
\hline 
 Mehta \etal \cite{mehta2017monocular}& 3DV\textquoteright 17  & 75.7  & 39.3  & 117.6\tabularnewline
 Mehta \etal \cite{mehta2017vnect}& ACM ToG\textquoteright 17  & 76.6  & 40.4  & 124.7\tabularnewline

Pavllo \etal \cite{pavllo20193d} ($f=81$) & CVPR\textquoteright 19  & 86.0  & 51.9  & 84.0\tabularnewline


Lin \etal \cite{lin2019trajectory} ($f=25$) & BMVC\textquoteright 19  & 83.6  & 51.4  & 79.8\tabularnewline

 Li \etal \cite{li2020cascaded}& CVPR\textquoteright 20  & 81.2  & 46.1 & 99.7\tabularnewline

 Chen \etal \cite{chen2021anatomy}& TCSVT\textquoteright 21  & 87.9  & 54.0  & 78.8\tabularnewline

 PoseFormer \cite{poseformer} (f = 9)& ICCV\textquoteright 21 & \underline{88.6}  & \underline{56.4} & \underline{77.1}\tabularnewline
\hline 
 CrossFormer (f=9)&  &  \textbf{89.1} & \textbf{57.5} & \textbf{76.3}\tabularnewline
 \hline
\end{tabular}
    \label{tab:INF}
\end{table}
\\
\textbf{Computational Complexity Analysis:}  Table \ref{table:complexity} shows the comparison with different methods of complexity analysis.
For the number of parameter analysis, it is relatively bigger than Poseformer in the three settings, it increases slightly. However, this increase is negligible in favor of accuracy gains. 
Apart from Poseformer, the number of parameters is still competitive to the other methods. It is also noticeable that the increase of frames does not translate in ab increase in the total number of parameters (only by hundreds). Regarding the FLOPs, the proposed method is not the best in comparison to the comparable methods. However, it only exhibits a slight increase over Poseformer. 
For frames per second (FPS), our method shows a slightly lower number compared to PoseFormer.
\vspace{-0.5em}
\begin{table}[t!]
\centering
\caption{Comparison between the proposed method and a set of previous methods in terms of the comparison are computational complexity, number of the parameters, MPJPE, and Frames Per Second (FPS). The experiments are conducted on Human3.6M under Protocol 1 with the  detected 2D pose
as input.}
    \centering
\begin{tabular}{l|cccccc}
\hline 
 & f  & Parameters (M)  & FLOPs (M)  & MPJPE &FPS \tabularnewline
\hline 
 Hossain and Little \cite{hossain2018exploiting}& - & 16.95  & 33.88  & 58.3 & -\tabularnewline
%
 Chen \etal \cite{chen2021anatomy}& 27  & 31.88 & 61.7  & 45.3 &410 \tabularnewline
 Chen \etal \cite{chen2021anatomy}& 81  & 45.53  & 88.9  & 44.6&315 \tabularnewline
 Chen \etal \cite{chen2021anatomy}& 243  & 59.18  & 116  & 44.1 &264\tabularnewline
 PoseFormer \cite{poseformer}& 9  & 9.58  & 150 & 49.9 & 320 \tabularnewline
 PoseFormer \cite{poseformer}& 27  & 9.59  & 452 & 47.0 & 297\tabularnewline
 PoseFormer \cite{poseformer}& 81  & 9.60 & 1358  & 44.3 &269\tabularnewline
\hline 
 CrossFormer &  9&  9.93& 163 &  48.5& 284\tabularnewline
 CrossFormer &  27&  9.93& 515 &46.5  & 266\tabularnewline
 CrossFormer &  81& 9.93 &1739  & 43.8 &241 \tabularnewline
\hline 
\end{tabular}
\centering
    \label{table:complexity}
\end{table}

\textbf{Qualitative Results.} In order to show the superiority of the proposed method qualitatively, we compare it with the ground-truth and Poseformer (the state-of-the art approach \cite{poseformer}). The evaluation is conducted on Human3.6 dataset S11 test set on ``SettingDown'' action. From Fig. \ref{fig:qualitative}, it is clear that the proposed method is considerably better than Poseformer. We use blue arrows to define the locations where our method behaves better. While our method shows some failures, it is  still overall better than Poseformer.
\subsection{Ablation Study}
In order to check the impact of the the proposed method individually, we perform empirical experiments on Human3.6 dataset using protocol 1. Also, one experiment is conducted on the optimal hyper-parameters selection.\\
\textbf{The impact of cross-joints modules} In this ablation study, we investigate the contribution of Cross-Joint Interaction (CJI) on the spatial transformer and on the whole network. We also study the impact of CJI with spatial embedding since it considers inductive bias implicitly and without. Regarding Cross-Frame Interaction (CFI) module, we verify its impact on the network independently with temporal embedding and without. For a fair comparison, we use the optimal parameter settings, including 4 blocks for both transformers. The dimension of the keypoints is unified to $32$ for the spatial and $544$ for the temporal. Table \ref{table:crossmodules} shows the results of various settings between CJI and CFI. The results illustrate that using both of the cross-joint modules improve the total performance significantly (from $49.9$ to $48.5$). Table \ref{table:crossmodules} (row 4) discusses using CJI module along with both embeddings and without CFI. It achieved better performance than Poseformer (from $49.9$ to $49.3$). This explains the need of both cross-joint modules to add locality to the transformers. Similarly, using CFI only along with the embeddings achieved better performance compared to PoseFormer. However, it achieved lower performance than CJI. It is clear that using cross-joint modules without the spatial information embeddings shows no difference in terms of performance, which proves our claim of using CJI and CFI modules to add to the locality of the self-attention modules. However, CJI is more independent than CFI to positional information as it shows the same accuracy without the positional information, whereas CFI obtains $0.04\%$~less.

Another ablation study (a table is included in the supplementary) is conducted to verify the importance of the spatial and temporal transformers. This shows the various settings of the proposed network and provides comparisons to Poseformer.

\begin{table}[t]
\centering
    \caption{Quantitative Comparison between Poseformer and the proposed method on the impact of different components on the total performance. The experiments are performed on Human3.6M (Protocol 1) using CPN 2D pose as input. Various settings are used for cross-joint modules (CJI: Cross-Joint Interactions; CFI: Cross-Frames Interactions; E$_s$: spatial embedding; E$_t$: temporal embedding)}
    \begin{tabular}{cccccc}
\hline 
frames & CJI & CFI &  E$_s$ & E$_t$ &   CrossFormer\tabularnewline
\hline 
9 &  \xmark& \xmark & \checkmark & \checkmark & 49.90  \tabularnewline
9 &  \checkmark& \xmark & \xmark & \xmark &49.30   \tabularnewline
9 & \checkmark& \xmark & \checkmark & \xmark & 49.30  \tabularnewline
9 & \checkmark& \xmark & \checkmark & \checkmark &  49.30 \tabularnewline
9 & \checkmark& \checkmark & \xmark & \checkmark &  48.50 \tabularnewline
9 &  \xmark&\checkmark  & \xmark &\xmark  &  49.83 \tabularnewline
9 &    \xmark&\checkmark  &\xmark  &\checkmark & 49.80 \tabularnewline
9 &    \xmark&\checkmark  &\checkmark  &\checkmark & 49.80 \tabularnewline
9 & \checkmark & \checkmark &  \checkmark& \checkmark  &48.50 \tabularnewline
\hline 
\end{tabular}
    \label{table:crossmodules}
\end{table}

\section{Conclusion}
Two interaction modules have been proposed to resolve the issues of using the spatial and temporal transformers for 3D human pose estimation. The first module, cross-joint interaction (CJI), has been presented to resolve the locality issue of the spatial transformers, while the second module, cross-frame interaction (CFI), has been developed to encode the dependencies of the joints across the subsequent frames. Both of two modules have been incorporated into transformer architecture, CrossFormer, and validated on popular 3D pose datasets. The proposed method has achieved new SOTA results for both datasets. In the future, we will test on other visual recognition applications to ensure its generalisation to different visual tasks.
{
\small
\bibliographystyle{ieee_fullname}
\bibliography{DB.bib}
}



\end{document}


\pagestyle{headings}
\mainmatter
\def\ECCVSubNumber{4193}  

\title{CrossFormer: Cross Spatio-Temporal Transformer for 3D Human Pose Estimation - Supplementary Material} 

\titlerunning{ECCV-22 submission ID \ECCVSubNumber} 
\authorrunning{ECCV-22 submission ID \ECCVSubNumber} 
\author{Anonymous ECCV submission}
\institute{Paper ID \ECCVSubNumber}

\maketitle

\section{Additional Ablation Study}
\subsection{The impact of Spatial-Temporal modules}\\
Additional to the experiments in Sec. $4.4$, we have implemented another ablation study to check the impact of including the spatial and temporal encoders within the same architecture. Table \ref{table:components} illustrates various settings of the proposed network and conducts comparison with Poseformer. Overall, the proposed method achieved superior performance compared to PoseFormer. More specifically, Using temporal transformer only does not show the best results. This is due to the need to encode the spatial relationships between the joints and the absence of the possibility to infer the locality within the architecture of the transformer. In contrast, using spatial only is not helping as the temporal relationships between body joints across frames are not preserved. Combining both spatial and temporal provides the best results. 
\begin{table}
    \centering
    \caption{Quantitative Comparison between Poseformer and the proposed method on the impact of different components on the overall performance. The experiments are performed on Human3.6M (Protocol 1) using CPN 2D pose as input. (T: Temporal only; S: Spatial only)}
\begin{tabular}{ccccc}
\hline 
frames & S & T  & Poseformer & CrossFormer\tabularnewline
\hline 
9 & \checkmark & \xmark & - & 55.24 \tabularnewline
9& \xmark & \checkmark & 52.5 & 51.37\tabularnewline
9 & \checkmark & \checkmark &  51.6 & 48.54 \tabularnewline

\hline 
\end{tabular}
    \label{table:components}
\end{table}

\subsection{Hyper-parameters Analysis}\\
In this study, we perform an analysis on various architecture settings to find the optimal hyper-parameters. As Table \ref{table:settings} shows, several settings have been chosen such as $32$ for embedding dimension, $4$ is the number of transformer blocks stacked on top of each other for both transformers. Overall, the optimal settings are $32$, $4$, $4$ for embedding dimension, spatial transformer blocks, temporal transformer blocks, respectively. It is clear that increasing the number of blocks degrades the performance. It is worth noting that our proposed method behaves better than Poseformer in all the settings by considerable margin. This reveals that the across-joints modules are advantageous  in all the scales.
\begin{table}[]
\centering
    \caption{Visual comparison between the proposed method and Poseformer as am ablation study on different architecture parameters. The experiment is conducted on Human3.6M (Protocol 1) using CPN 2D pose as input. d is the dimension of the spatial transformer. L$_s$ refers to the number of
transformer blocks the spatial transformer, where as  L$_t$  for  temporal transformers. All the experiments use 9 as the number of frames.}
    \begin{tabular}{ccccc}
\hline 
 d& L$_s$ & L$_t$ & Poseformer & CrossFormer  \tabularnewline
\hline 

16 & 4 & 4 & 51.7 & 50.6  \tabularnewline

32 & 4 & 4 & 49.9 & 48.5  \tabularnewline

32 & 6 & 6 & 50.3 & 49.2  \tabularnewline

48 & 4 & 4 & 50.4 & 49.7  \tabularnewline

48 & 6 & 6 & 50.5 & 50.1  \tabularnewline
64 & 4 & 4 & - & 50.6  \tabularnewline
64 & 6 & 6 & - & 50.9  \tabularnewline

\hline 
\end{tabular}
    \label{table:settings}
\end{table}

\begin{table}[]
\caption{Quantitative results using Mask-rcnn (MRCNN) detections and ground-truth (GT) on HumanEva-I dataset under
protocol #2. (\textcolor{red}{Red}: best)}
\resizebox{0.97\columnwidth}{!}{\begin{tabular}{l|ccc|ccc|ccc|c}
\hline 
\multirow{2}{*}{} & \multicolumn{3}{c|}{Walk} & \multicolumn{3}{c|}{Jog} & \multicolumn{3}{c|}{Box} & \multirow{2}{*}{Avg}\tabularnewline
\cline{2-10} \cline{3-10} \cline{4-10} \cline{5-10} \cline{6-10} \cline{7-10} \cline{8-10} \cline{9-10}   
 & S1 & S2 & S3 & S1 & S2 & S3 & S1 & S2 & S3 & \tabularnewline
\hline 
 Martinez \etal \cite{martinez2017simple}& 19.7 & 17.4 & 46.8 & 26.9 & 18.2 & 18.6 & - & - & - & -\tabularnewline
 Pavlakos \etal \cite{pavlakos2018ordinal}& 22.3 & 19.5 & \textcolor{red}{29.7} & 28.9 & 21.9 & 23.8 & - & - & - & -\tabularnewline
Lee \etal \cite{lee2018propagating} & 18.6 & 19.9 & 30.5 & 25.7 & 16.8 & 17.7 & 42.8 & 48.1 & 53.4 & 30.3\tabularnewline
Pavllo \etal \cite{pavllo20193d} & 13.9 & 10.2 & 46.6 & 20.9 & 13.1 & 13.8 & 23.8 & 33.7 & 32.0 & 23.1\tabularnewline
Li \etal \cite{li2021exploiting} (f = 27 MRCNN)& 14.0 & 10.0 & 32.8 & \textcolor{red}{19.5} & 13.6 & 14.2 & 22.4 & 21.6 & 22.5 & 18.9\tabularnewline
CrossFormer (f = 27 MRCNN)  &\textcolor{red}{13.5}  &\textcolor{red}{9.2}  & 32.9 & 21.8 &\textcolor{red}{12.5}  &\textcolor{red}{12.6}  & \textcolor{red}{21.2} & \textcolor{red}{21.2} & \textcolor{red}{21.8} &\textcolor{red}{18.5}\tabularnewline
\hline
Li \etal \cite{li2021exploiting} (f = 27 GT) & 9.7 & 7.6 & 15.8 & 12.3 & 9.4 & 11.2 & 14.8 & 12.9 & 16.5 & 12.2\tabularnewline
 CrossFormer (f = 27 GT)&  \textcolor{red}{8.2} &  \textcolor{red}{6.8} &  \textcolor{red}{14.3} &  \textcolor{red}{11.1} &  \textcolor{red}{8.4} &  \textcolor{red}{10.5} &  \textcolor{red}{13.2} &  \textcolor{red}{11.7} &  \textcolor{red}{15.0} &  \textcolor{red}{11.0}\tabularnewline
\hline 
\end{tabular}}
\label{tab:my_label}
\end{table}

{
\small
\bibliographystyle{ieee_fullname}
\bibliography{DB.bib}
}